# Symbolic Generalization for On-line Planning

**Zhengzhu Feng**
Computer Science Department
University of Massachusetts
Amherst, MA 01003
fengzz@cs.umass.edu

**Eric A. Hansen**
Department of Computer Science and Engineering
Mississippi State University
Mississippi State, MS 39762
hansen@cse.msstate.edu

**Shlomo Zilberstein**
Computer Science Department
University of Massachusetts
Amherst, MA 01003
shlomo@cs.umass.edu

## Abstract

Symbolic representations have been used successfully in off-line planning algorithms for Markov decision processes. We show that they can also improve the performance of on-line planners. In addition to reducing computation time, symbolic generalization can reduce the amount of costly real-world interactions required for convergence. We introduce Symbolic Real-Time Dynamic Programming (or sRTDP), an extension of RTDP. After each step of on-line interaction with an environment, sRTDP uses symbolic model-checking techniques to generalizes its experience by updating a group of states rather than a single state. We examine two heuristic approaches to dynamic grouping of states and show that they accelerate the planning process significantly in terms of both CPU time and the number of steps of interaction with the environment.

## 1 Introduction

Markov decision processes (MDPs) have been adopted as a framework for research in decision-theoretic planning. Classic dynamic programming algorithms solve MDPs in time polynomial in the size of the state space. However, the size of the state space grows exponentially with the number of features describing the problem. This "state explosion" problem limits use of the MDP framework, and overcoming it has become an important topic of research.

Over the past several years, symbolic representations have been used successfully to improve the performance of off-line planning algorithms for MDPs. For example, Dearden & Boutilier (1997) proposed a feature-based (or factored) representation of MDPs that uses decision trees as a compact representation. The SPUDD algorithm (Hoey *et al.* 1999) achieved improved performance using a decision diagram based representation, adapted from the symbolic model-checking community. Feng & Hansen (2002) combined SPUDD with the LAO* algorithm, as a way of integrating state abstraction with heuristic search. These approaches focus on how to perform off-line planning (via dynamic programming) more efficiently. In this paper, we introduce a symbolic generalization of Real-Time Dynamic Programming (RTDP) (Barto, Bradtke, & Singh 1995), an on-line planner for MDPs. We call this algorithm symbolic RTDP, or sRTDP. Whereas RTDP uses an on-line state trajectory to focus computation and determine what individual states to backup, sRTDP uses an on-line state trajectory to determine what abstract states to backup. That is, sRTDP generalizes experience using state abstraction.

The ability to generalize experience is crucial for online algorithms such as RTDP, both when the state space is large and when obtaining experience is relatively expensive compared to the cost of computation. The key issue in generalization is the identification of "similar" states. Previous work has focused on generalization based on *input similarity*, as measured by some distance metric defined over the representation space of the states. However, as pointed out by Yee (1992), input similarity does not necessarily lead to similarity in the underlying value function of the MDP, limiting the effectiveness of this approach. In this paper, we propose to generalize experience based on *structural similarity*, capturing better the underlying value function. States are considered similar if they have similar value estimates, or similar reachability structures. We argue that structural similarity is a more effective approach to generalization because the value estimates and the reachability structure are directly related to the underlying value function of the MDP. Symbolic model-checking techniques are particularly useful in this approach to generalization, because they enable us to efficiently identify structural similarity without enumerating the state space.

## 2 Background

We begin with a brief review of MDPs and algorithms for solving MDPs, including value iteration, LAO* and RTDP. Then we review factored MDPs and methods of state abstraction that use decision diagrams.

### 2.1 Markov Decision Processes

A Markov decision process (MDP) is defined as a tuple $M = (S, A, P, R)$ where: $S$ is a set of states; $A$ is a set of actions; $P$ is a set of transition models of the form $P^a : S \times S \rightarrow [0,1]$, where $P^a(s, s')$ is the probability of making a transition from state $s$ to state $s'$ if action $a$ is taken in state $s$; and $R$ is a set of reward models of the form $R^a : S \rightarrow \Re$, where $R^a(s)$ is the expected reward for taking action $a$ in state $s$. We consider MDPs for which the objective is to find a policy $\pi : S \rightarrow A$ that maximizes total discounted reward over an infinite (or indefinite) horizon, where $\gamma \in [0,1]$ is the discount factor. (We allow a discount factor of 1 only for MDPs that reach a terminal state, i.e., zero-reward absorbing state, with probability 1.)

Starting with an arbitrary state evaluation function $V^0 : S \rightarrow \Re$, the standard dynamic programming (DP) algorithm updates the value function at every state $s$ as follows:

$$V^{t+1}(s) \leftarrow \max_{a \in A} \left\{ R^a(s) + \gamma \sum_{s' \in S} P^a(s, s') V^t(s') \right\}. \tag{1}$$

Value Iteration(VI) solves an MDP by successively applying this DP update, and the sequence of value functions converges to the optimal value function $V^*$ in the limit (Puterman 1994). The optimal policy $\pi^* : S \rightarrow A$ can be obtained from $V^*$ by setting each $\pi^*(s)$ equal to the action that maximizes the right-hand side of Equation (1) when $V^t = V^*$.

Note that the standard DP update is performed on every state in the state space. This is not necessary if the agent is given some starting state(s) and only a part of the state space is reachable from there. The algorithms LAO* (Hansen & Zilberstein 2001) and RTDP (Barto, Bradtke, & Singh 1995) exploit this fact by limiting the DP update to a subset of the state space. They differ mainly in the way this subset is determined. LAO* is an off-line algorithm that performs best-first search in the state space. It interleaves a forward step that expands the current policy to find reachable states, and a backward step that performs a DP update on the found states. RTDP is an on-line algorithm that interacts directly with the environment (or a simulation of it), and performs updates on states that are actually visited in the course of interaction. Both algorithms can solve an MDP without necessarily visiting the whole state space, and converge to a solution that is optimal for all relevant states.

### 2.2 Factored MDPs and Decision Diagrams

In a factored MDP, the set of states is described by a set of random variables $\mathbf{X} = \{X_1, \ldots, X_n\}$. Without loss of generality, we assume these are Boolean variables. Using $x_i$ to denote an instantiation of a state variable $X_i$, a particular instantiation of the variables corresponds to a unique state, denoted $s = \{x_1, \ldots, x_n\}$. Because the size of the state space grows exponentially with the number of variables, it is impractical to represent the transition and reward models explicitly as matrices when the number of states variables is large.

To achieve a compact representation, we use decision diagrams (Bryant 1986; Bahar *et al.* 1993). Algebraic decision diagrams (ADDs) are a generalization of binary decision diagrams (BDDs), a compact data structure for Boolean functions that is used in symbolic model checking. A decision diagram is a data structure (corresponding to an acyclic directed graph) that compactly represents a mapping from a set of Boolean state variables to a set of values. A BDD represents a mapping to the values 0 or 1. An ADD represents a mapping to any finite set of values. To represent these mappings compactly, decision diagrams exploit the fact that many instantiations of the state variables map to the same value. In other words, decision diagrams exploit state abstraction. BDDs are typically used to represent the characteristic functions of sets of states and the transition functions of finite-state automata. ADDs can represent weighted finite-state automata, where the weights correspond to transition probabilities or rewards, and thus are an ideal representation for MDPs.

The SPUDD algorithm (Hoey *et al.* 1999) was the first to use the above representation in solving MDPs. Let $\mathbf{X} = \{X_1, \ldots, X_n\}$ represent the state variables at the current time and let $\mathbf{X}' = \{X'_1, \ldots, X'_n\}$ represent the state variables at the next step. For each action $a$ and each post-action variable $X'$, an ADD $P^a(\mathbf{X}, X')$ represents the probability that $X'$ becomes *true* after action $a$ is taken. The complete action ADD $P^a(\mathbf{X}, \mathbf{X}')$ can be computed by multiplying the ADDs for each variable (Hoey *et al.* 1999). Similarly, the reward model $R^a(\mathbf{X})$ for each action $a$ is represented by an ADD. The advantage of using ADDs to represent mappings from states (and state transitions) to values is that the complexity of operators on ADDs depends on the size of the diagrams, not the size of the state space. If there is sufficient regularity in the model, ADDs can be very compact, allowing problems with large state spaces to be represented and solved efficiently.

SPUDD implements the standard DP update as follow:

$$V^{t+1}(\mathbf{X}) \leftarrow \max_{a \in A} \left\{ R^a(\mathbf{X}) + \gamma \exists_{\mathbf{X}'} P^a(\mathbf{X}, \mathbf{X}') \cdot V^t(\mathbf{X}') \right\}.$$

Note that the value functions $V^t$ and $V^{t+1}$ are represented using ADDs, and all operators involved in the DP update are applied to ADDs. In particular, $\exists$ denotes the *existential abstraction* operator, which sums over all post-action states. We refer to (Hoey *et al.* 1999) for detailed discussion and related references. Compared to traditional DP using a tabular representation, SPUDD exploits state abstraction by *implicitly* grouping states with the same value into an abstract state, and performing computation on the abstract state space. We say "implicitly" because these abstract states are never singled out during the computation. Instead, the symbolic operators automatically take advantage of abstraction found in the ADD representation.

Symbolic LAO* (Feng & Hansen 2002) is an extension of LAO* that uses the same representation as SPUDD. Like LAO*, it interleaves a forward search step that expands the current policy and constructs the set of reachable states, denoted $E$, with a DP step that updates the values of states in $E$. The forward step is implemented as a form of symbolic reachability analysis, a common operation in symbolic model checking. The set $E$ is represented by its characteristic function $\chi_E$ using an ADD. The DP update is a modified version of the SPUDD algorithm that uses the following *masked* update to focus computation on the relevant part of the state space:

$$V_E^{t+1}(\mathbf{X}) \leftarrow \max_{a \in A} \left\{ R_E^a(\mathbf{X}) + \gamma \exists_{E'} P_{E \cup E'}^a(\mathbf{X}, \mathbf{X}') \cdot V_{E'}^t(\mathbf{X}') \right\} \quad (2)$$

Here $E'$ is the set of states reachable from $E$. The notation $f_E(\cdot)$ stands for the "masked" version of ADD $f$, which is the product of $f$ and the characteristic function of $E$: $f_E = f \times \chi_E$. The operation of masking constrains the DP update to a subset of the state space, and is primarily responsible for the performance improvement of symbolic LAO* over SPUDD. Symbolic LAO* also performs better than LAO* because it exploits state abstraction in both the forward search and DP steps.

## 3 Symbolic RTDP

Recall that RTDP performs a DP update while interacting with the environment. At each time step $t$, the agent observes the current state $s_t$ and performs a DP backup to update its value, as follows:

$$V^{t+1}(s_t) \leftarrow \max_{a \in A} \left\{ R^a(s_t) + \gamma \sum_{s' \in S} P^a(s_t, s') V^t(s') \right\}. \quad (3)$$

The values of all other states are kept unchanged, that is, for all $s \neq s_t$:

$$V^{t+1}(s) = V^t(s).$$

If the initial value function is an admissible heuristic estimate of the optimal value function, then always taking the action that maximizes Equation (3) results in convergence to an optimal value function. Otherwise some exploration scheme must be used in choosing actions, in order to ensure convergence. After an action is taken, the agent observes the resulting state and the cycle repeats.

An advantage of RTDP over standard DP is that it uses an on-line trajectory of states, beginning from the start state, to determine which states to update, and as a result, unreachable states are not updated. However, the enumerative nature of the trajectory sampling makes it difficult to scale up to large state spaces. When the state space is very large, a state-by-state update becomes inefficient, especially if the sampling involves carrying out physical actions.

We now describe *symbolic RTDP*, or sRTDP, a symbolic version of RTDP that helps overcome this inefficiency by generalizing the update from a single state to an abstract state. Figure 1 shows the pseudo-code of a trial-based version of sRTDP. It takes as input an admissible initial value function $V_0$, a starting state $s_0$, the number of trials to run, and the number of steps to run in each trial. It returns an updated value function, from which a policy can be extracted.

We extend the idea of masking from symbolic LAO* to sRTDP by performing DP on the abstract state $E$ that the current state $s$ belongs to. Symbolic model-checking provides us with convenient and efficient techniques to group states as abstract states and to manipulate these abstract states. There are many ways to group states into abstract states. In this paper, we examine two heuristic approaches that are motivated by the idea of generalization by structural similarity. A *value-based* abstract state consists of states whose value estimates are close to that of the current state. A *reachability-based* abstract state consists of states that share with the current state a similar set of successor states. Unlike SPUDD, we *explicitly* construct this abstract state at each time step of sRTDP, using standard operations on ADDs. We use the function *Generalize*$(s)$ for this operation in Figure 1. The two heuristic approaches to implementing *Generalize*$()$ are described below:

**sRTDP**($V_0, s_0, nTrials, nSteps$)
1. $V \leftarrow V_0$;
2. Repeat $nTrials$ times
3. $s \leftarrow s_0$;
4. Repeat $nSteps$ times
5. $E \leftarrow Generalize(s)$
6. $E' \leftarrow$ States reachable in one step from $E$
7. $V^{copy} \leftarrow V$
8. For all $a \in A$:
9. $Q_a \leftarrow R^a_E(\mathbf{X}) + \gamma \exists_{E'} P^a_{E \cup E'}(\mathbf{X}, \mathbf{X}') \cdot V_{E'}(\mathbf{X}')$
10. $V_E \leftarrow \max_{a \in A} Q_a$
11. $a \leftarrow \arg\max_{a \in A} Q_a(s)$
12. $V \leftarrow V_E + V^{copy}_{\bar{E}}$
13. $s \leftarrow Execute(s, a)$
14. Return $V$

Figure 1: Trial-based sRTDP algorithm

**Generalization by Value** With a value-based abstract state, the experience is generalized to states that have similar value estimates as the current state. Generalizing updates to states with the same or similar *estimated* values helps the agent in two ways. First, if some of these states indeed have a similar optimal value as the current state, the update strengthens this similarity and the agent is better informed in the future when these states are visited again. On the other hand, if some of the states have very different optimal values than the current state, the generalization helps to distinguish them and their values are not recomputed when the current state is visited again.

Let $s$ be the current state and let $V$ be the current value function. The characteristic function of the value-based abstract state $E$ can be constructed by setting leaf nodes in $V$ with values close to $V(s)$ to 1, and all other leaf nodes to 0. The change at the leaf nodes then propagates up to the root. This operation is standard in most ADD packages, including CUDD (Somenzi 1998), the one we use for our implementation.

**Generalization by Reachability** With a reachability-based abstract state, experience is generalized to states that are similar to the current state in terms of the set of one-step reachable states. The intuition is that if the agent is going to visit some states, say $C$, from the current state $s$, then any information about $C$ is useful not only to $s$ but also to other states that can reach $C$. By generalizing the update to these other states, the agent is better informed in the future about whether to aim at $C$ or to avoid it.

To compute the abstract state based on reachability, we introduce two operators from the model-checking literature. The $Img(C)$ operator computes the set of one-step reachable states from states in $C$, and the $PreImg(C)$ operator computes the set of states that can reach some state in $C$ in one step. The reachability-based abstract state $E$ can then be computed as:

$$E = PreImg(Img(\{s\})) - PreImg(S - Img(\{s\})).$$

Once the abstract state $E$ is identified, we use Equation (2) to update its value. Since all elements on the right-hand side of the equation are masked, the resulting ADD on the left hand side is effectively masked by $E$ also (hence the $V_E$ notation on the left hand side of Equation (2) and line 10 of the algorithm). In line 12, we merge this masked value function back to the whole state space in order to obtain an updated value function. The $\bar{E}$ notation stands for the complement of $E$. After the update, an action is chosen that maximizes the DP update at state $s$. The agent then carries out the action, denoted $Execute(s, a)$, and the process repeats.

Although both symbolic LAO* and sRTDP use a *masked* DP update, the masks they use are different and serve different purposes. The mask in symbolic LAO* contains all states visited so far by the forward search step. The purpose of masking is to restrict computation to relevant states. The mask in sRTDP contains states that share structural similarity. The purpose of masking is to generalize the update of a single state to an abstract state. This generalization has two consequences. First, it introduces some overhead for identifying the abstract state, and for performing masked DP instead of single-state DP. On the other hand, it updates the value of a group of states in a single step, at a cost that can be significantly less than updating the states individually. For problems that have a large state space but regular structure, the benefit of masking can be much greater than its overhead.

**Convergence** If we implement the function *Generalize*($s$) so that it only returns the set $\{s\}$, then sRTDP becomes RTDP. On a state-by-state level, the only difference between RTDP and sRTDP is that RTDP updates the current state only, while sRTDP updates the current state *and* some other states. Thus, if the convergence conditions for RTDP are met, sRTDP will also converge as long as the current state is always updated.

**Theorem 1** *sRTDP converges to the optimal value function under the same conditions that RTDP converges if for every state* $s$, $s \in Generalizes(s)$.

## 4 Adaptive sRTDP

Barto, Bradtke, & Singh (1995) describe an adaptive version of RTDP where the model parameters are not known and have to be estimated on-line while the agent is acting. It is straightforward to extend sRTDP to this setting. We call this algorithm *adaptive sRTDP*, or AsRTDP. Learning algorithms developed for Bayesian networks can be applied to learn the model parameters of a factored MDP, for example (Friedman & Goldszmidt 1999; Saul & Jordan 1999). To create an adaptive version of sRTDP, we modify the algorithm in Figure 1 to use the learned model in Equation (2). The identification of the abstract state remains the same. To satisfy the convergence conditions for adaptive RTDP, we use a simple $\epsilon$-greedy exploration scheme to replace the action selection step at line 11 of the algorithm. Finally, since there is no model to begin with, it is generally not possible to compute an admissible heuristic (although a good initial estimate of the value function can still speed up convergence).

## 5 Experimental Results

In this section, we consider the empirical performance of sRTDP and AsRTDP, and the performance of the two methods of generalization. We compare their performance to symbolic LAO*, RTDP and an adaptive version of RTDP. In our comparison, all algorithms use the same symbolic representation of the problem. Non-symbolic RTDP uses a symbolic representation because our test problems are too large for a traditional table-based representation of the transition matrix to fit in memory. However, non-symbolic RTDP performs single-state DP backups using Equation (3) in our comparison, and does not exploit the symbolic representation in solving the MDP.

We tested the various algorithms on the same test problems used in (Feng & Hansen 2002), especially the most difficult of these problems, numbered *a1* through *a4*. These four problems are adapted from the widget-processing problem used in (Hoey *et al.* 1999), with the modification that every state variable is affected by at least one action, and actions have different, random rewards. The results for these problems are very similar and we only report results for problem *a1* here. It has 20 Boolean state variables and 25 actions.

### 5.1 Symbolic RTDP

We first compare the performance of sRTDP, using generalization by value and by reachability, with symbolic LAO* and non-symbolic RTDP. The on-line planning algorithms performed 100 trials, each consisting of 20 steps from the starting state. The off-line planner, symbolic LAO*, ran until convergence. All used the same admissible heuristic function as an initial value function. (We used the easily-computed heuristic function $\frac{r^{max}\gamma}{1-\gamma}$, where $r^{max}$ denotes the maximum one-step reward.)

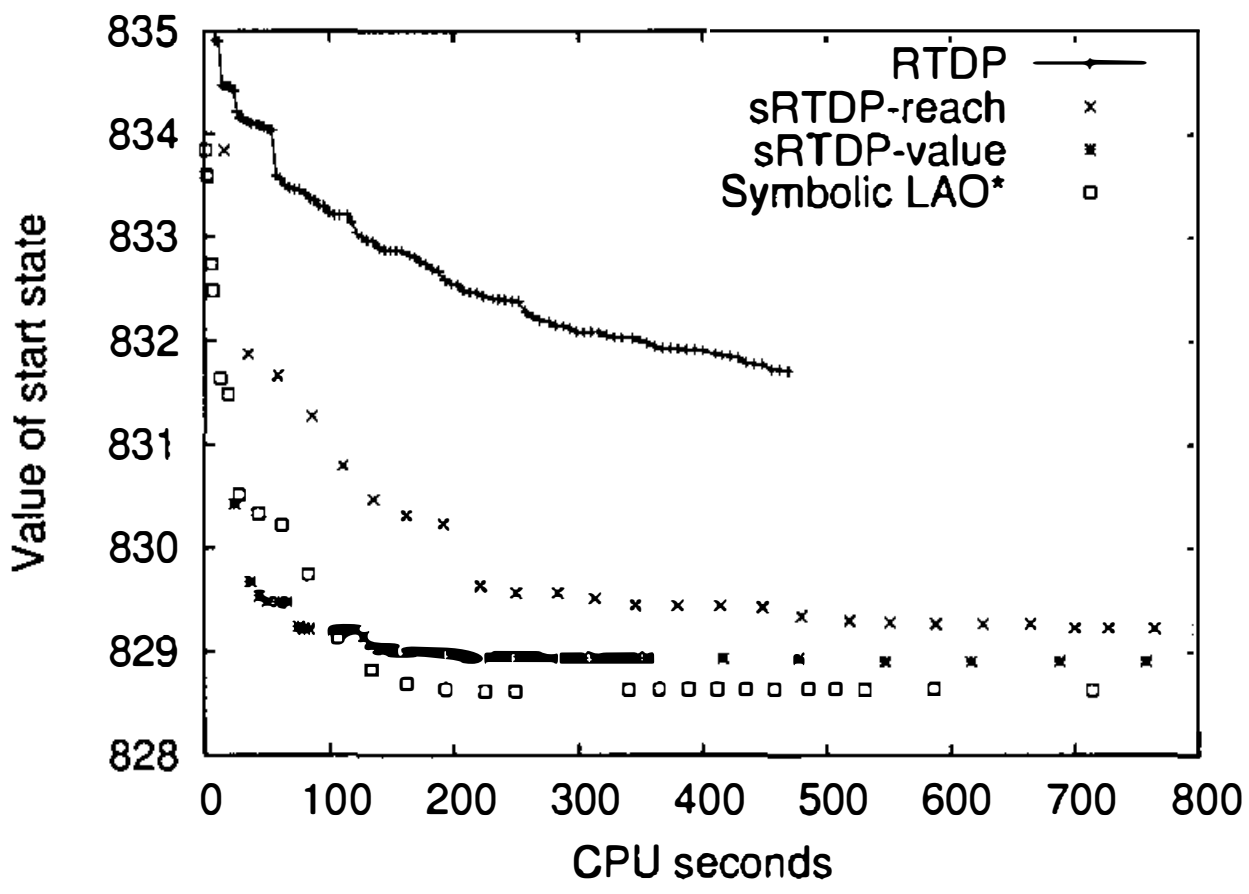

Figure 2: Performance comparison of non-adaptive algorithms in terms of CPU time.

The result is shown in Figure 2. The $x$-axis shows CPU time measured in seconds. The $y$-axis shows the value of the start state, which all algorithms attempt to optimize. Each point on the symbolic LAO* curve represents an iteration of forward search, followed by a DP update. Each point on the three RTDP curves represents a trial of 20 steps. As we can see, the two sRTDP algorithms perform much better than RTDP. This is because sRTDP generalizes experience and exploits state abstraction, while RTDP does not. sRTDP also compares favorably with symbolic LAO*. In particular, sRTDP with generalization by value quickly reaches a near-optimal value in the early stage of computation, while symbolic LAO* gradually catches up after about 100 seconds. Symbolic LAO* converges after running about 8 minutes, while sRTDP continues without reaching the same value even at the end of the 100 trials. This behavior – in which sRTDP improves a solution more quickly at first, and symbolic LAO* achieves eventual convergence faster – is similar to behavior observed in comparing non-symbolic versions of LAO* and RTDP (Hansen & Zilberstein 2001). The explanation is that RTDP focuses on high-probability paths, which results in early improvement, whereas LAO* considers all reachable states equally.

From Figure 2, we can also see that sRTDP takes longer to finish each trial than RTDP. In fact, RTDP finishes 100 trials in about 500 seconds, while the two sRTDP algorithms only finish from 20 to 40 trials in the same time. However, in each trial sRTDP improves the value function more than RTDP. If we plot the RTDP curves against the number of trials, shown in Figure 3, the difference becomes more obvious. After about 20 trials, sRTDP reaches a value that is within 0.1 of the value that symbolic LAO* converges to. For RTDP, the difference in value is larger than 2.1 after 100 trials. Since RTDP updates a single state only at each step, it takes less time to finish a trial than sRTDP, which performs the extra work of identifying and updating the abstract state at each step. However, the extra work by sRTDP leads to improved performance due to state abstraction and generalization.

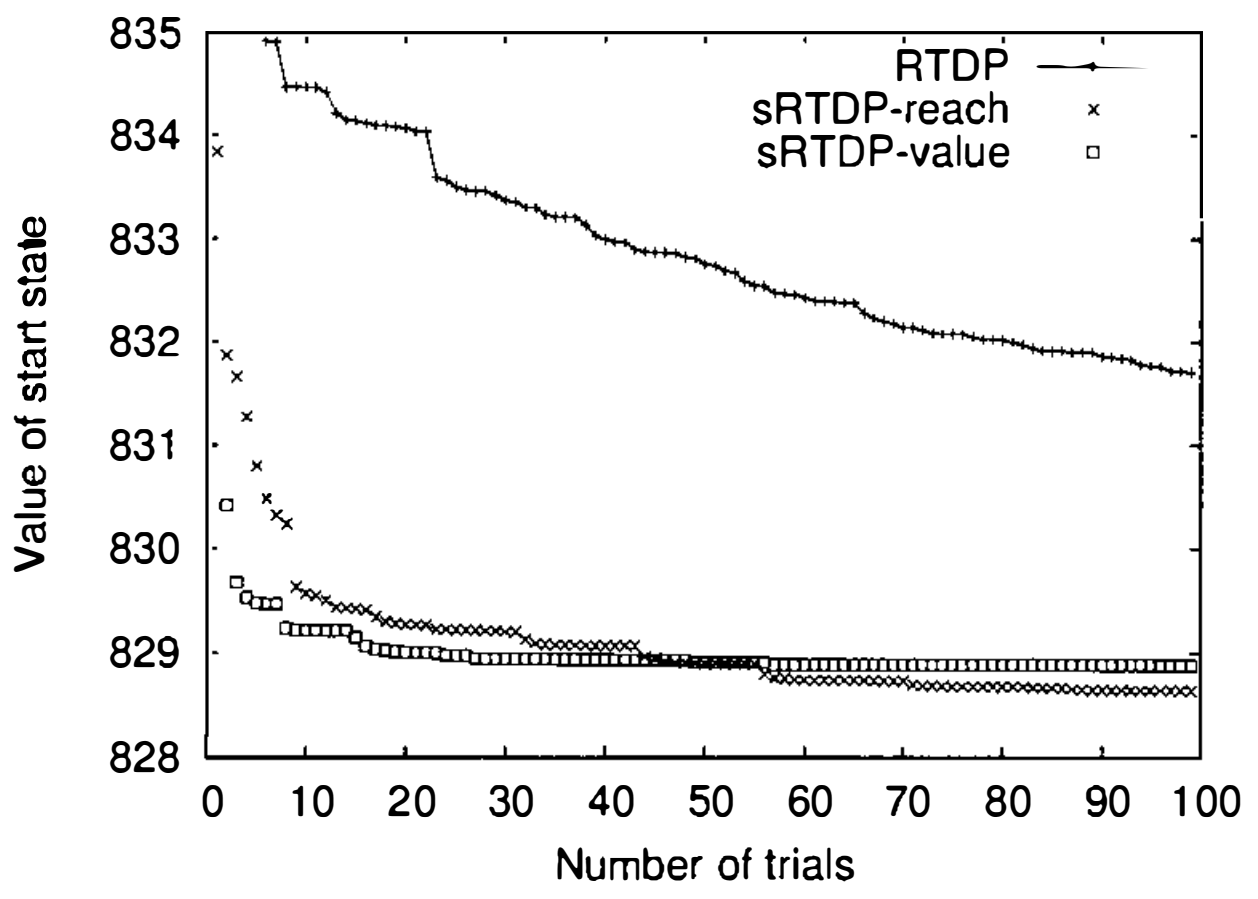

Figure 3: Performance comparison of non-adaptive versions of RTDP in terms of number of trials.

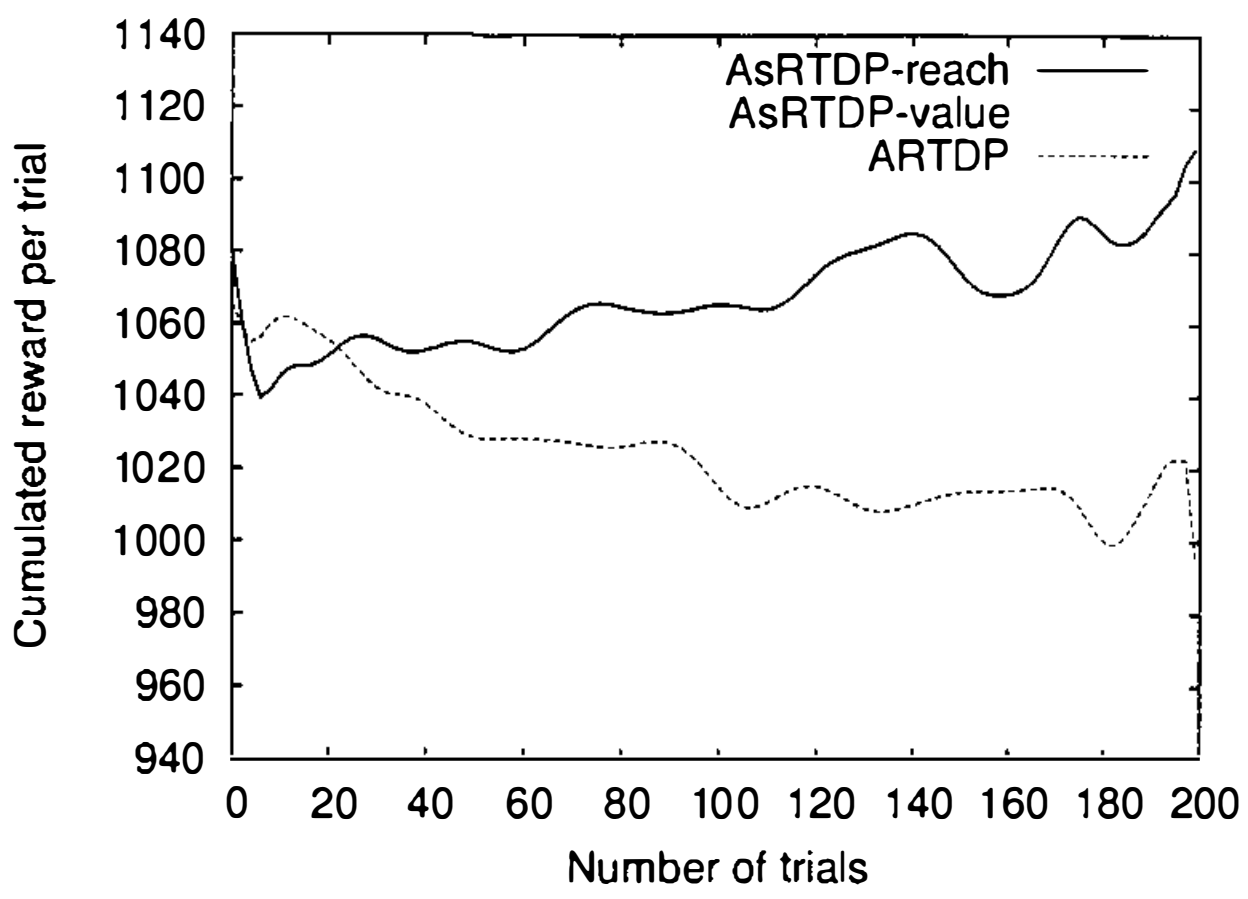

Figure 4: Performance comparison of adaptive versions of RTDP, averaged over 100 runs and smoothed.

From Figures 2 and 3, we can see that the two notions of generalization work similarly well for this problem, with generalization by value slightly better than generalization by reachability. We expect that the relative performance of the two methods will depend on the characteristics of a problem. In particular, if the current value estimation is close to the underlying optimal value function, as is the case when an admissible heuristic is used, value-based generalization should work better. Otherwise reachability-based generalization can be more effective, as we will see next.

### 5.2 Adaptive sRTDP

We next compare adaptive versions of sRTDP that use the two generalization approaches, with an adaptive version of non-symbolic RTDP. Since model learning is not the focus of this paper, we introduce two assumptions for this task to simplify the implementation: (1) the reward function is given; and (2) the structure of the transition ADD $P^a(\mathbf{X}, X')$ is given for all actions $a$ and state variables $X'$. Given these assumptions, we use a simple maximum-likelihood algorithm to estimate the missing probabilities. Since an admissible heuristic cannot be computed without an accurate model, we set the initial value function to 0 for our experiments.

Figure 4 shows the results. Each curve represents the accumulated reward in each trial, and is averaged over 100 runs and smoothed. Each run contains 200 trials with 20 steps per trial. As we can see, the two AsRTDP algorithms consistently outperform adaptive RTDP (or ARTDP). Moreover, while we see a clear trend that the AsRTDP curves are improving, the ARTDP curve seems to show no improvement over time. This is because AsRTDP generalizes its on-line experience, while ARTDP does not. Recall that the problem has 20 state variables, or 1,048,576 states. Each run performs $200 \times 20 = 4,000$ times of sampling, which is less than 0.4% of the state space. (Since some states may be sampled more than once, the actual sample coverage is likely to be smaller.) Since ARTDP does not generalize, sample coverage at this magnitude is far from enough. AsRTDP, on the other hand, generalizes beyond the actual samples, and is able to improve its performance based on the same amount of experience available to ARTDP.

By comparing the two sRTDP curves, we can see that generalization by reachability performs better than generalization by value. In fact, generalization by value has the worst on-line performance among the three algorithms over the first 60 trials. This is because in the early stage, the value estimates are very inaccurate, so the computation performed by generalization by value is mainly geared toward distinguishing states that have similar estimates but indeed have different optimal values. As experience accumulates, the value estimates become more accurate and generalization by value can better exploit it to gather more reward. This suggests a mixed strategy that applies different forms of generalization at different stages of the trials. We leave this to future work.

## 6 Related Work

Like RTDP, *prioritized sweeping* (PS) (Moore & Atkeson 1993) interleaves planning (and learning) with on-line interaction with an MDP. After performing a backup of the current state, PS performs additional backups of other states before taking its next action. It uses a priority queue to select additional states to backup in an order that reflects the likelihood of improvement, based on propagating changed values of states to predecessor states. Because it updates multiple states after each action, PS accelerates convergence and can reduce the amount of on-line interaction needed. In this respect, PS is similar to sRTDP, although the two algorithms choose additional states to backup in different ways.

The original PS algorithm performs value and priority updates on a state-by-state basis, without exploiting problem structure, and this can cause significant overhead. As a result, PS has been generalized to use symbolic representations. Andre, Friedman, & Parr (1998) describe *generalized prioritized sweeping*, which uses a parametric representation of an MDP to generalize model updates to similar states and adjust priorities accordingly. Dearden (2001) describes *structured prioritized sweeping*, which uses a compact, decision-tree representation of the value function and exploits state abstraction in value updates by using a *local decision-theoretic regression* operator that is closely related to the masking operator described in this paper. Structured PS differs from sRTDP in that it generalizes a backup to states with similar priority, whereas sRTDP generalizes a backup to states with similar value or reachability structure. Use of a priority queue also implies multiple updates after each action, whereas sRTDP performs a single symbolic update. (Other differences between PS and RTDP may affect the selection of states to update. In particular, RTDP focuses computation on states that are reachable from a specific starting state.)

The idea of extending the backup of a single state to an abstract state is closely related to function approximation methods for solving MDPs. Neural networks, for example, are often used to represent a value function compactly using a relatively small number of parameters (Bertsekas & Tsitsiklis 1996). Because a DP update improves the value function by adjusting these parameters, a small change can affect the values of a group of states or even the whole state space. As a result, some states may get updated as a result of the approximation mechanism instead of from dynamic programming. This makes it difficult to analyze the convergence properties of such algorithms. In fact, it has been shown that function approximation methods can sometimes diverge, or converge to a value function that is arbitrary bad in quality (Boyan & Moore 1995). In contrast, sRTDP guarantees convergence to optimality because the symbolic representation we use is an exact representation. But it is also worth mentioning that our representation does not exclude the possibility of approximation. By grouping similar but not identical state values together, we can reduce the size of the ADDs and the DP update can be computed more efficiently. This form of approximation has been studied for standard DP algorithms (St-Aubin, Hoey, & Boutilier 2001; Feng & Hansen 2001) and shown to converge with bounded error. A similar approach to approximation may also be adapted for use with sRTDP.

Our work is also related to the idea of model minimization for MDPs, presented in (Dean & Givan 1997). Their *model minimization* algorithm constructs a *stochastic bisimulation* (Larsen & Skou 1991) for a factored MDP. The bisimulation consists of abstract states that are equivalent in terms of optimal value and optimal policy. A potentially smaller MDP is constructed over this abstract state space and the optimal solution for it is also optimal for the original MDP. Our algorithm can be roughly viewed as an on-line version of model minimization (Yannakakis & Lee 1993), interleaved with an update of the value function using dynamic programming. The benefit of on-line model minimization is that unreachable states are not distinguished, so that a potentially much smaller abstract state space is traversed than in full MDP model minimization. By interleaving DP updates with model minimization, we also don't have to wait until the minimal model is created before performing value updates.

## 7 Conclusion

Generalization has long been recognized as a crucial component of efficient planning and learning. It accelerates the learning process and reduces the amount of interaction with the environment needed to reach a desired level of competence. We have described sRTDP, an extension of RTDP that uses symbolic model-checking techniques as an approach to generalizing experience in solving factored MDPs. By identifying and updating abstract states instead of single states, sRTDP improves a state evaluation function faster than RTDP not only in terms of CPU time, but also in terms of the number of steps of interaction with the environment. This is particularly desirable when performing real-world actions is more expensive than performing computation, which is the case in many applications. The result is a novel generalization technique for on-line planning that accelerates convergence without compromising optimality.

### Acknowledgements

We thank the anonymous reviewers for helpful comments. Support for this work was provided in part by the National Science Foundation under grants IIS-0219606 and IIS-9984952 and by NASA under Cooperative Agreement NCC-2-1311. Zhengzhu Feng was supported by a UMass Graduate School Fellowship.